# Using Contact to Increase Robot Performance for Glovebox D&D Tasks

Aykut Özgün Önol, Philip Long, Taşkın Padır*

*Robotics and Intelligent Vehicles Research (RIVeR) Lab, Northeastern University, Boston, MA.

**ABSTRACT**

Glovebox decommissioning tasks usually require manipulating relatively heavy objects in a highly constrained environment. Thus, contact with the surroundings becomes inevitable. In order to allow the robot to interact with the environment in a natural way, we present a contact-implicit motion planning framework. This framework enables the system, without the specification in advance of a contact plan, to make and break contacts to maintain stability while performing a manipulation task. In this method, we use linear complementarity constraints to model rigid body contacts and find a locally optimal solution for joint displacements and magnitudes of support forces. Then, joint torques are calculated such that the support forces have the highest priority. We evaluate our framework in a 2.5D, quasi-static simulation in which a humanoid robot with planar arms manipulates a heavy object. Our results suggest that the proposed method provides the robot with the ability to balance itself by generating support forces on the environment while simultaneously performing the manipulation task.

**INTRODUCTION**

As nuclear facilities reach the end of their life cycle they must be decommissioned in a safe and efficient manner. A particularly dangerous task is the decontamination of gloveboxes that have been previously used to manipulate radioactive material. In Savannah River Site (SRS) in South Carolina alone there are more than 3500 gloveboxes in use today. Each glovebox task requires at least three workers: an operator, a radiological control inspector, and a supervisor. In spite of strictly enforced safety procedures, accidents can occur; for instance in 2010, a technician received a puncture wound while using a glovebox and the accident resulted in internal contamination with transuranic elements [1].

During the decommissioning task, the radioactive material is airborne accentuating the already extreme hazards faced by the operator. In addition to high risk levels and strenuous working conditions the task itself is composed of dull repetitive motions. For instance, a decommissioning task requires transporting all debris and objects from the interior of the glovebox to the exit port. Although a robotic system that is specifically designed for glovebox operations may be the best solution, humanoid robots are an attractive option since they can operate in a variety of environments and use tools that are designed for humans. With these motivations, our objective is to evaluate the technological readiness level of dexterous robot hands, such as those on NASA's humanoid robot Valkyrie [2] to replace human hands for safe and risk averse operations in existing gloveboxes in nuclear facilities. Due to the critical nature of the work, the system is human supervised nevertheless the robot must still possess a high degree of autonomy to cope with un-modeled surroundings and dynamic events.

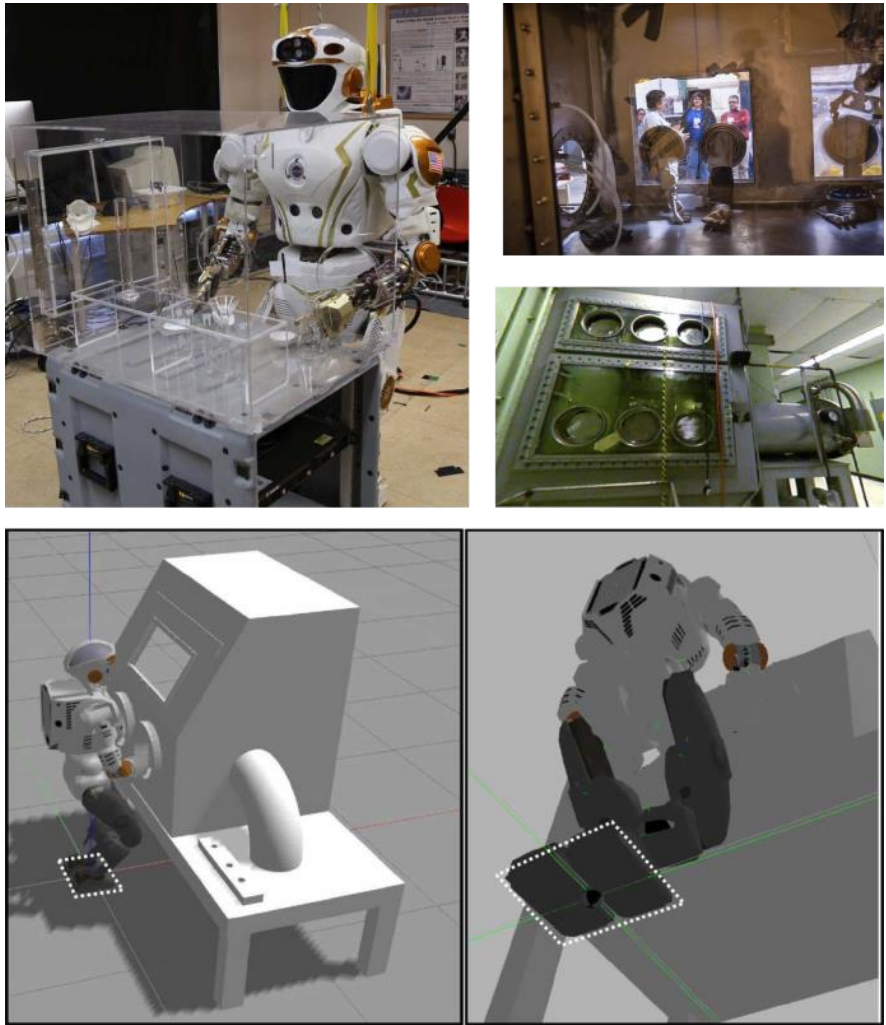

Fig. 1. Representation of glovebox manipulation task. Top left, Valkyrie robot interacting with objects in prototype glovebox environment. Top right, actual gloveboxes in use in nuclear facilities. Bottom dynamic simulation of Valkyrie interacting with glovebox model, with support polygon outline.

In order to conduct operations within the glovebox the constraints imposed by the ports, gloves and the external structure, as shown in Fig. 1, must be considered. In particular, as Valkyrie is manipulating objects inside the glovebox, the forearms are effectively fixed at the entry ports. The inability to alter body configuration greatly diminishes the robot's capacity to take steps in arbitrary directions. This in turn leads to a real danger of toppling during task execution as the system cannot easily change the support polygon's location. One possible solution would be to use advanced motion planning and control algorithms [3-5] to avoid contact with the ports while maintaining stability. Sugihara and Nakamura [6] analyze the stability and the arm manipulability for a two dimensional humanoid robot and show the relation between the degree of manipulability and the stability. In [7] the case of pushing light objects with a humanoid robot is investigated and its consequence on the zero-moment point (ZMP) is illustrated. However, while manipulating heavy objects such as tools and jigs, a typical requirement for decommissioning, it could be challenging to find feasible motions while maintaining the robot's equilibrium.

Toppling occurs when the zero-moment point (ZMP) [8] leaves the support polygon (SP). If the support polygon cannot be displaced, alternative methods to maintain stability must be employed. For example in [9] the ZMP position for a cable driven mobile robot is modified on-line by a tension distribution algorithm. In [10], an increase in support polygon size by using supplementary contact points is shown. In contrast to these works, we propose to exploit the contacts in the glovebox (i.e., leaning on the entry ports) in order to shift the ZMP towards the center of the support polygon while performing manipulation tasks.

In this study, rigid body contacts are modeled through linear complementarity constraints (LCCs), and a contact-implicit motion planning method based on nonlinear constrained optimization is presented. In order to mitigate potential numerical difficulties, the equality constraint in LCCs is relaxed into an inequality constraint. In the optimization procedure, we solve for joint displacements, the magnitude of support forces, and a slack variable that is used for the relaxation such that the position error for the object being manipulated, control effort, and the slack variable associated with the relaxation are minimized. This process is subject to constraints that ensure the object is grasped by the end effectors, the ZMP is in a safe region, and the deviation of the object's position from a desired position is admissible. Furthermore, we use a multi-objective torque controller that primarily generates the support forces obtained from optimization and projects the torques needed for object wrench into the null space of support forces. The proposed methodology is tested through 2.5D, quasi-static simulations by assuming a humanoid robot with two planar arms manipulating a relatively heavy object on an elevated plane representing the glovebox.

The remainder of the paper is organized as follows. An overview of relevant work is given in Related Work. The optimization process, motion planning and torque controller is outlined in Methodology. The simulation experiments and results are described in Results. Finally, the conclusions are drawn and avenues of future work are discussed in Conclusion.

**Related Work**

In order to achieve contact-rich behaviors, for instance locomotion and manipulation in cluttered environments, both the contact modes (i.e., contact postures and contact transitions) and the constrained dynamics must to be taken into account. One approach is to use a contact-before-motion planner such as those presented in [11,12]. In this case, first a sequence of contacts at predefined locations is determined, then the trajectories are obtained subject to contact constraints. In contrast if a motion-before-contact planner is used the contacts are a result of the motion trajectories [11]. Alternatively a complementarity-based contact models can be used in a constrained optimization framework. The use of complementarity constraints to model rigid body contacts with friction was first proposed in [13,14]. In [15], the quasi-static motion of a set of planar rigid bodies in contact is modeled through complementarity constraints, and bilinear programming is used to predict its motion. Recently, such models have been used in time-stepping integration schemes for trajectory optimization through contact (or contact-implicit trajectory optimization) for locomotion of legged robots and manipulation with simple grippers [16-18]. The main idea here is to consider the contact-mode-related parameters as additional optimization variables so that the contact modes need not be determined beforehand [17]. Such an optimization problem are shown to be solved locally through gradient-based constrained optimization algorithms such as sequential quadratic programming [14,19].

When the desired robot configuration and contact forces are obtained, the torques required to generate these forces can be calculated through a null-space-based multi-objective control approach. Park and Khatib [20] proposed a torque control framework for humanoid robots with multiple contacts and verified the method experimentally in [21]. Moreover, they extended this method to a unified hierarchical whole-body control framework for humanoid robots in [22]. In this framework, tasks are hierarchically ranked. Thus the torques required for a lower-priority task are projected into the null-space of the Jacobian matrix associated with a higher-priority task. In [23], a whole-body torque controller for humanoid robots is proposed that combines passivity-based balancing proposed in [24] with a hierarchical null-space-based control that is similar to [22].

## METHODOLOGY

### Static Equilibrium

In order for a robot to be balanced, it needs to be in static equilibrium [8]. In this case, for the static equilibrium of the system, the following wrenches needs to be considered: the wrench due to the robot's mass, the wrenches at the end effectors due to the object wrench, and the wrenches at the support contact points. Henceforward, we enumerate the left and right arms as the first and second arms, respectively.

The static equilibrium for forces can be written as:

$$\sum \mathbf{f} = 0 \Rightarrow \mathbf{f}_R + m\mathbf{g} + \sum_{i=1}^{2}\mathbf{f}_{Si} + \sum_{i=1}^{2}\mathbf{f}_{Ci} = \mathbf{0} \qquad \text{(Eq. 1)}$$

where $m$ is the total mass of the robot, $\mathbf{g}$ denotes the gravity vector given by $\mathbf{g} = [0,0,-9.81]^T$. $\mathbf{f}_{Si}$ is the force at the support points between the second link of the $i$-th arm and the boundaries of the glovebox port, $\mathbf{f}_{Ci}$ denotes the force at the contact point between the object and the end effector of the $i$-th arm, and $\mathbf{f}_R$ is the ground reaction force.

Additionally, the projection of the moment onto the horizontal plane (i.e., *xy*-plane in this case) must be zero, i.e., $M_x = 0$ and $M_y = 0$:

$$\sum \mathbf{M}_O^H = \mathbf{0} \Rightarrow \left[\mathbf{p}_R \times \mathbf{f}_R + \mathbf{p}_{CoM} \times m\mathbf{g} + \sum_{i=1}^{2}\left(\mathbf{p}_{Ci} \times \mathbf{f}_{Ci} + \mathbf{M}_{Si} + \mathbf{p}_{Ci} \times \mathbf{f}_{Ci} + \mathbf{M}_{Ci}\right)\right]^H = \mathbf{0} \qquad \text{(Eq. 2)}$$

where $\mathbf{a}^H$ denotes the vector containing the *x* and *y* or *horizontal* components of a vector $\mathbf{a}$. $\mathbf{p}_R$, $\mathbf{p}_{CoM}$, $\mathbf{p}_{Si}$, and $\mathbf{p}_{Ci}$ are the vectors denoting the position of the ground reaction force (whose *x* and *y* components represent the ZMP), the robot's center of mass (CoM), the support points on the glovebox ports and the contact points on the object, respectively. $\mathbf{M}_{Si}$ denotes the moment applied at the support contact point by the *i*-th arm, while $\mathbf{M}_{Ci}$ denotes the moment applied on the object by the *i*-th arm. The position of the ZMP, $\mathbf{p}_R$, is obtained by solving Eqs. (1) and (2) simultaneously. In order to avoid toppling, the ZMP must lie in the support polygon (SP), namely, the convex hull of the robot's feet.

The object wrench $\mathbf{h}_o \in \mathbb{R}^6$ can be obtained in terms of the wrenches applied by the end effectors as follows [25]:

$$\mathbf{h}_o = \begin{bmatrix} \mathbf{W}_{C_1} & \mathbf{W}_{C_2} \end{bmatrix} \begin{bmatrix} \mathbf{h}_{C_1} \\ \mathbf{h}_{C_2} \end{bmatrix} = \mathbf{W}_C \mathbf{h}_C \qquad \text{(Eq. 3)}$$

where $\mathbf{W}_{Ci} \in \mathbb{R}^{6\times 6}$ is the wrench matrix that transforms the wrench at the $i$-th contact point, $\mathbf{h}_{Ci} \in \mathbb{R}^6$, to the wrench at the origin of the object frame, which is the center the object in this case, and given by:

$$\mathbf{W}_{Ci} = \begin{bmatrix} \mathbf{I}_3 & \mathbf{0}_3 \\ -\hat{\mathbf{r}}_{Ci} & \mathbf{I}_3 \end{bmatrix} \quad \text{(Eq. 4)}$$

where $\hat{\mathbf{a}}$ is the skew-symmetric matrix representation of the vector $\mathbf{a}$, $\mathbf{r}_{Ci}$ is the vector from the *i*-th contact point $\mathbf{p}_{Ci}$ to the origin of the object frame, $\mathbf{I}_3$ is 3×3 identity matrix, and $\mathbf{0}_3$ is 3×3 zero matrix. Then, given the object wrench, the wrenches at the end effectors can be calculated from $\mathbf{h}_C = \mathbf{W}^+\mathbf{h}_O$, where $\mathbf{W}^+$ is the Moore-Penrose pseudo-inverse of the matrix $\mathbf{W}$.

**Motion Planning**

In this work, we ignore dynamic effects and investigate the quasi-static case for dual-arm manipulation of a relatively heavy object in a glovebox (i.e., a very confined space). In the following robot's joint position and velocity are denoted by $\theta$ and $\dot{\theta}$, while those of the *i*-th arm are referred to as $\theta_i$ and $\dot{\theta}_i$. The objective is to preserve the robot's balance during the manipulation task. In other words our goal is to find the configurations that would keep the robot's ZMP in the safe region by leaning on the glovebox ports while simultaneously maintaining the manipulated object's desired position. For this purpose, we use direct optimization to find the joint displacements and support contact forces in a contact-implicit manner, namely, without planning for contact points beforehand.

In order to take into account the rigid body contacts, we use the complementarity constraints that are given by [14,17,18] as:

$$\begin{aligned} \phi(\mathbf{q}) &\geq \mathbf{0}, \\ \gamma &\geq \mathbf{0}, \\ \gamma^T \phi(\mathbf{q}) &= 0 \end{aligned} \quad \text{(Eq. 5)}$$

where $\mathbf{q}$ is the generalized configuration vector, $\phi(\mathbf{q})$ is the signed distance vector between closest points on all pairs of bodies (a robot's link and a candidate contact point in the environment). $\gamma$ is the Lagrange multiplier that corresponds to the vector of magnitudes of the support forces in the normal direction. The first constraint prevents any interpenetration, the second ensures that the bodies can only push each other, and the last allows force generation only when bodies are in contact. Thus, only one of these variables (either $\phi(\mathbf{q})$ or $\gamma$) can be non-zero. We relax the equality constraint by converting it into an inequality constraint through a slack variable. After which, as described in [18,19], the slack variable is added into the cost function to minimize the relaxation so that potential numerical issues are mitigated.

As a result, the following optimization problem is solved to find the joint displacements $\delta\theta = \theta_{k+1} - \theta_k$, the magnitude of the normal contact forces $\gamma$, and the slack variable $s$:

$$\underset{\delta\theta,\gamma,s}{\text{minimize}} \ w_1 \left\| \mathbf{p}_o^d - \mathbf{p}_o \right\|^2 + w_2 \left\| \delta\theta \right\|^2 + w_3 s$$

subject to

$$g(\mathbf{q}) = 0,$$
$$\gamma, s \geq 0, \quad \quad \quad \quad \quad \quad \quad \quad \quad \quad \quad \quad \text{(Eq. 6)}$$
$$s - \gamma^T \phi(\mathbf{q}) \geq 0,$$
$$r_s - \left\| \mathbf{p}_R^d - \mathbf{p}_R \right\| \geq 0,$$
$$r_o - \left\| \mathbf{p}_o^d - \mathbf{p}_o \right\| \geq 0$$

where $g(\mathbf{q}) = 0$ ensures the end effectors are in contact with the object, $w_i$ is the weight (a positive scalar) associated with the $i$-th term of the cost function, $\|\cdot\|$ is the Euclidean norm, $\mathbf{p}_R^d$ is the desired position of the ZMP (i.e., the center of the SP), $\mathbf{p}_o$ and $\mathbf{p}_o^d$ are the object's actual and desired positions respectively. Finally, $r_s$ and $r_o$ are the radii of the safe circle (SC) for the ZMP and the allowable deviation of the object position, respectively.

A local solution for such a nonlinear programming problem can be found by gradient-based nonlinear constrained optimization algorithms such as interior-point and SQP. In this work, we use the SQP, as in [17], since it has been shown to be generally more advantageous for optimal control problems than the interior-point algorithm [26].

**Torque Control**

Using the optimization procedure, described in the previous section, the robot configuration and the support forces' magnitude are obtained. Torque control is required to generate object acceleration and also to generate the support forces necessary to maintain robot equilibrium.

The torques necessary to generate the desired object wrench $\tau_h$ can be obtained as:

$$\tau_h = \begin{bmatrix} \mathbf{J}_1^T & \mathbf{0} \\ \mathbf{0} & \mathbf{J}_2^T \end{bmatrix} \mathbf{W}_C^+ \mathbf{h}_o = \mathbf{J}^T \mathbf{W}^+ \mathbf{h}_o \quad \quad \text{(Eq. 7)}$$

$\mathbf{J}_i$ is the kinematic Jacobian matrix that relates the joint velocities of the $i$-th arm to the translational velocities at the end effector $\mathbf{v}_i$, i.e., $\mathbf{v}_i = \mathbf{J}_i \dot{\boldsymbol{\theta}}_i$.

The support forces are oriented normal to the contacting robot link, thus using the contact angle $\beta_i$ and the resulting force $\gamma$, the support force that is in contact with the $i$-th arm can be calculated as:

$$\mathbf{f}_{Si} = \frac{1}{2} \begin{bmatrix} \gamma_i \cos(\beta_i) & \gamma_i \sin(\beta_i) & 0 \end{bmatrix}^T \quad \quad \text{(Eq. 8)}$$

Similarly, the joint torques required to generate these forces, denoted as $\tau_S$, can be calculated as described in [21]. In our case, there are a maximum of two supports points at any instant, therefore

$$\tau_S = \begin{bmatrix} \mathbf{J}_{S_1}^T & \mathbf{0} \\ \mathbf{0} & \mathbf{J}_{S_2}^T \end{bmatrix} \begin{bmatrix} \mathbf{f}_{S_1} \\ \mathbf{f}_{S_2} \end{bmatrix} = \mathbf{J}_S^T \mathbf{f}_S \quad \text{(Eq. 9)}$$

$\mathbf{J}_{Si}$ is the Jacobian matrix that relates the joint velocities of the $i$-th arm to the translational velocities at the corresponding support point.

For the glovebox task, the support forces are crucial to maintain the stability of the robot, while generating the desired object wrench has a lower priority. Thus, we project $\tau_h$ into the null space of the Jacobian matrix for support forces $\mathbf{J}_S$ to obtain the overall joint torques $\tau$, as:

$$\tau = \tau_S + \mathbf{N}_S \tau_h \quad \text{(Eq. 10)}$$

where

$$\mathbf{N}_S = \mathbf{I}_{n_d} - \mathbf{J}_S^T (\mathbf{J}_S^T)^+ \quad \text{(Eq. 11)}$$

is the null space projector of the support forces, and $n_d$ is the number of degrees of freedom (DOF) of the whole robot. Consequently, the resulting joint torques would generate the desired support forces to ensure the balance of the robot and create an object wrench using the redundancy of the robot.

## RESULTS & DISCUSSION

In order to test the proposed framework, we run simulation experiments in which a humanoid robot that has two planar 4-DOF arms with revolute joints (i.e., 8 DOF in total) manipulates a relatively heavy rigid bar on an elevated plane. The robot's arms pass through two ports representing the glovebox, as shown in Fig. 2. Furthermore, we omit velocities and accelerations (i.e., quasi-static case) and assume point contacts without friction. The values of the weights are selected as follows $w_1 = 10^3$, $w_2 = 10^2$, and $w_3 = 10^6$ by trial and error such that the optimization yields feasible results. Initial values for all of the optimization variables are zero. The radii of tolerance circles for the ZMP and the object position ($r_s$ and $r_o$) are selected as 0.15 m and 0.1 m, respectively. Finally, the masses of the robot and the object are 54 kg and 12 kg, respectively, and the desired motion of the object in simulations is in $+y$-direction; therefore, the desired object wrench to generate an acceleration in this direction is given as $\mathbf{h}_o = [0,10,-117.72,0,0,0]^T$.

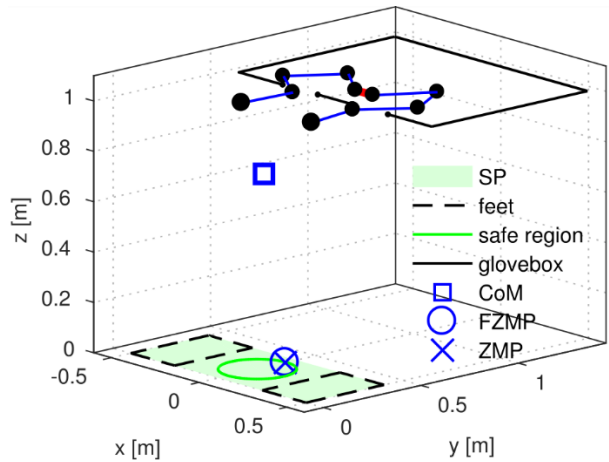

Fig. 2. Simulation environment for dual-arm manipulation in a glovebox.

We investigate the task of moving the object 40 cm forward on a straight path that consists of 9 equally spaced way-points. The results are depicted in Fig. 3. Each step of the motion is indicated by a color from blue to red. In the initial configuration (indicated by blue), the robot grasps the object from both ends.

During the simulation, the position of the ZMP is calculated with and without the effect of the support contacts on the glovebox frame. The latter is known as the fictitious ZMP (FZMP) [8] since it may fall outside of the SP.

The results show that the contact-implicit motion planning method performs successfully. The robot makes contacts with the glovebox to maintain its balance while moving the object on the desired path. As soon as the FZMP leaves the SR in the second step, the right arm makes a contact with the left end of the port to push the ZMP into the SR. As the object moves further away from the base, the contact angle is varied so that the magnitude of the support force in $-x$-direction is larger. This is required due to the circular shape of the SR. However as object moves further from the base, simply changing the contact angle is no longer sufficient, thus the left arm also makes contact with the right end of the left port. As a result, the object is successfully transported along the desired straight path with a position error of 0.1 m (i.e., the allowed deviation) in each step after the initial configuration.

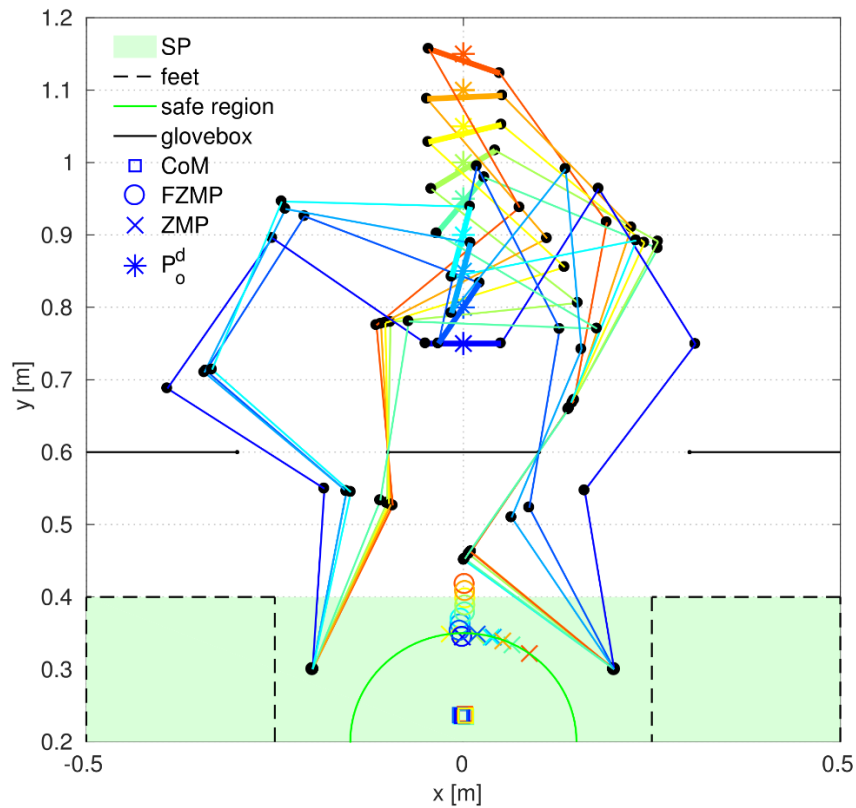

Fig. 3. Simulation results while moving the object on a straight path.

An enlarged version of the SP area is illustrated in Fig. 4 to show the change of the ZMP and the FZMP throughout the simulation. It can be seen that the FZMP moves forward along with the object's position. Nonetheless, the ZMP does not leave the SR owing to the support forces generated by the motion planner. Furthermore, in step 7 (indicated by yellow), the ZMP is more centralized compared to the previous and the next steps due to the symmetry of the support forces with respect to the $y$-axis.

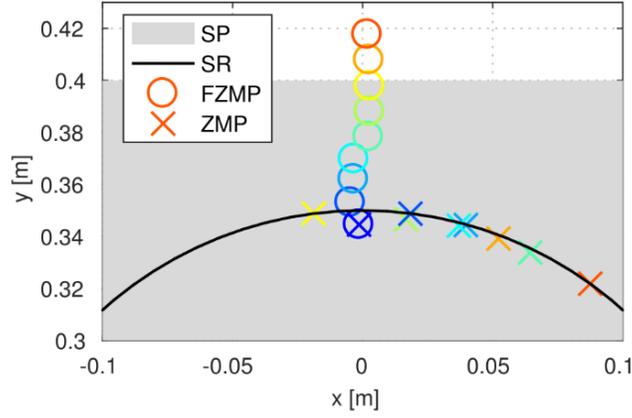

Fig. 4. Change of FZMP and ZMP throughout the simulation.

Figures 5(a) and (b) show the magnitudes of the support forces and the joint torques on the arms, respectively, with respect to the distance from the object to the robot's base. The magnitude of support forces are much bigger than the magnitude of the object wrench so that the torques are much more affected by the support forces than the object wrench. Thus, force and torque vs. distance characteristics are quite similar -- i.e., the torque is dominated by the support forces (especially after step 5). Moreover, the changes of $\|\mathbf{f}_S\|$ and $\|\mathbf{\tau}\|$ with the distance are almost linear. Aside from this, the magnitude of the support force on each arm is quite similar to each other in step 7, as consistent with the observation regarding the more centralized ZMP in step 7. Except for steps 1 and 7, the magnitude of the support force on the right arm is always bigger than the one on the left arm, which shifts the ZMP in +$x$-direction. Such an unbalanced distribution of forces might be undesirable due to the fact that higher joint torque limits would be required. Thus, enforcing a more uniform distribution of the support forces may be a future work.

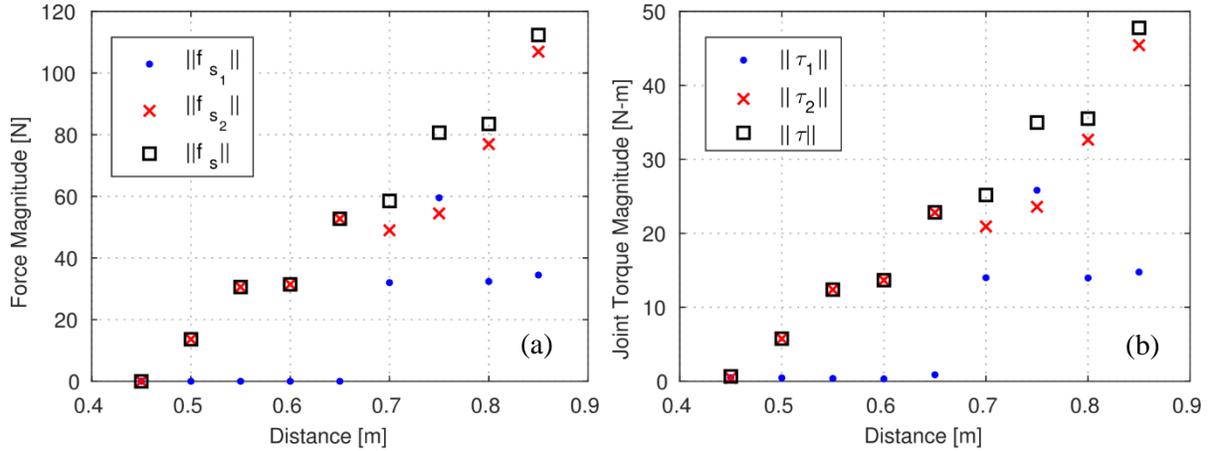

Fig. 5. (a) Magnitude of support forces vs. the distance of the object from the base, and (b) magnitude of joint torques vs. the distance of the object from the base.

## CONCLUSION

In this paper, we examined the scenario in which a humanoid robot performs dual-arm manipulation tasks in a glovebox. Glovebox operations require the displacement of heavy objects in a confined space, where taking steps in arbitrary directions to balance the system is not possible. Thus, we propose a contact-implicit motion planning framework that allows the robot to make and break contacts with the environment without a priori specification of contact modes. By using this system the robot's balance is maintained during the execution of a manipulation task. In other words, this motion planning method yields locally optimal joint displacements and support forces so that the robot can perform the task without falling over.

Once the optimization process has obtained a suitable joint configuration, a joint torques are calculated to firstly satisfy high priority task constraints, i.e. generate sufficient support forces, and secondly to generate a desired object wrench. 2.5D quasi-static simulations are used to evaluate the proposed framework in which a humanoid robot with two planar arms manipulates a relatively heavy object on an elevated plane. Our results demonstrate that the robot is able to interact with the environment to balance itself and carry out the manipulation task successfully.

Future work will focus on modeling the systems dynamics and the experimental validation of this algorithm using the Valkyrie humanoid robot.

## ACKNOWLEDGEMENT

This material is based upon work supported by the Department of Energy under Award Number DE-EM0004482, by the National Aeronautics and Space Administration under Grant No. NNX16AC48A issued through the Science and Technology Mission Directorate and by the National Science Foundation under Award No. 1451427.